\documentclass[runningheads,anonymous]{llncs}
\usepackage{amsmath,graphicx,multicol,multirow,booktabs}
\usepackage{bbm}
\usepackage[colorlinks,linkcolor=black,anchorcolor=black, citecolor=black]{hyperref}
\usepackage{amssymb}
\usepackage{enumitem}
\begin{document}
\title{Using Structural Similarity and Kolmogorov-Arnold Networks for Anatomical Embedding of Cortical Folding Patterns}
\titlerunning{Using Structural Similarity and KAN for Anatomical Embedding of 3HG}
%
\author{Minheng Chen \inst{1} \and Chao Cao \inst{1}\and Tong Chen \inst{1} \and Yan Zhuang \inst{1}\and Jing Zhang \inst{1}\and Yanjun Lyu\inst{1} \and Xiaowei Yu\inst{1}\and Lu Zhang \inst{2} \and Tianming Liu \inst{3} \and Dajiang Zhu\inst{1}}
%
\authorrunning{ M. Chen et al.}
%
\institute{ Department of Computer Science and Engineering, University of Texas at Arlington, USA \\
\email{\{dajiang.zhu\}@uta.edu}\and
Department of Computer Science, Indiana University Indianapolis, USA
\and
School of Computing,  University of Georgia, USA
}
\maketitle              
\begin{abstract}
The 3-hinge gyrus (3HG) is a newly defined folding pattern, which is the conjunction of gyri coming from three directions in cortical folding. Many studies demonstrated that 3HGs can be reliable nodes when constructing brain networks or connectome since they simultaneously possess commonality and individuality across different individual brains and populations.
However, 3HGs are identified and validated within individual spaces, making it difficult to directly serve as the brain network nodes due to the absence of cross-subject correspondence. The 3HG correspondences represent the intrinsic regulation of brain organizational architecture, traditional image-based registration methods tend to fail because individual anatomical properties need to be fully respected. To address this challenge, we propose a novel self-supervised framework for anatomical feature embedding of the 3HGs to build the correspondences among different brains. The core component of this framework is to construct a structural similarity-enhanced multi-hop feature encoding strategy based on the recently developed Kolmogorov-Arnold network (KAN) for anatomical feature embedding. Extensive experiments suggest that our approach can effectively establish robust cross-subject correspondences when no one-to-one mapping exists.

\end{abstract}

\section{Introduction}
\label{sec:intro}
The cortical folding patterns contain underlying mechanisms of brain organization.
While the major cortical regions are largely consistent across individuals, the local shapes and patterns within these regions vary significantly, posing challenges for the quantitative and efficient characterization of cortical folding.
Recently, a finer scale cortical folding pattern, known as the 3-hinge gyrus (3HG)~\cite{chen2017gyral,li2010gyral}, has been identified and defined as the junction where three gyri converge from different directions.
Notably, the 3HG is evolutionarily conserved across multiple primate species and remains stable in the human brain~\cite{li2017commonly,zhang2018exploring}, regardless of population differences or brain conditions. 
This cortical folding pattern exhibits strong consistency within species while varying among individuals, with unique features such as the thickest corteces~\cite{li2010gyral}, higher DTI-derived fibers density~\cite{ge2018denser},  and greater connectivity diversity across structural and functional domains compared with other gyral regions~\cite{zhang2020cortical}. This suggests that 3HGs are more like hubs in the cortical-cortical connection network and play a vital role in the global structural and functional networks of humans~\cite{liu2022optimized}. 
And a recent study~\cite{lyu20204mild} revealed that a finer-scale brain connectome based on 3HG can better capture the intricate  patterns of Alzheimer's Disease.

3HG-based brain networks analysis shows great potential in advancing our understanding of human brain and brain diseases. However, 3HGs are identified in individual space and there is no natural cross-subject correspondences. Therefore, identifying cross-subject correspondences of 3HGs is a key step for constructing 3HG networks and conducting population-wide analysis ~\cite{cao2024enhance,zhang2020identifying,zhang2024learning}. 
However, this process is highly challenging due to substantial individual variability.
In our previous work, \textit{cortex2vector}~\cite{zhang2023cortex2vector}, we introduced a learning-based embedding framework to encode individual cortical folding patterns as a series of anatomically meaningful embedding vectors. The similarities of the embedding vectors can effectively measure the correspondence between corresponding 3HGs.
This method is capable of establishing anatomical correspondences across individuals, as well as between different brains at various stages of neurodevelopment~\cite{zhang2024learning}.
However, the existing framework has several shortcomings:
1) the adopted multi-hop feature encoding strategy fails to fully capture the topological information of the folding patterns, as it only considers node proximity~\cite{wang2016structural} without accounting for the structural similarity of the nodes themselves; 
2) it remains uncertain whether the simple 4-layer multi-layer perceptron network can adequately learn the features sufficient to identify correspondences between different 3-hinge nodes with similar topological structures;
3) the sparsity of the topological feature network increases the likelihood that our framework may reconstruct insignificant zero elements within the features, thereby diminishing its representational capacity.

In this paper, to address the aforementioned limitations,
we propose a self-supervised framework for anatomical feature embedding of the 3HG based on our initial study~\cite{zhang2023cortex2vector}.
We introduce structural similarity between independent nodes to enhance the hierarchical multi-hop encoding strategy.
To further improve the representation ability of the network  while keeping it lightweight, we adopt Kolmogorov–Arnold Networks (KAN)~\cite{liu2024kan}, a recently proposed neural network inspired by the Kolmogorov-Arnold representation theorem, for anatomical feature encoding of 3HGs. In addition, we propose a new loss function -- \textit{selective reconstruction loss}, which penalizes reconstruction errors in non-zero elements, thereby enhancing the representational capacity of the embedding vector.
Our experimental results show that the learned embeddings can accurately establish cross-subject correspondences in complex cortical landscapes, while also maintaining the commonality and variability inherent in 3HGs.
\begin{figure}[htb]

  \centering
  \centerline{\includegraphics[width=\linewidth]{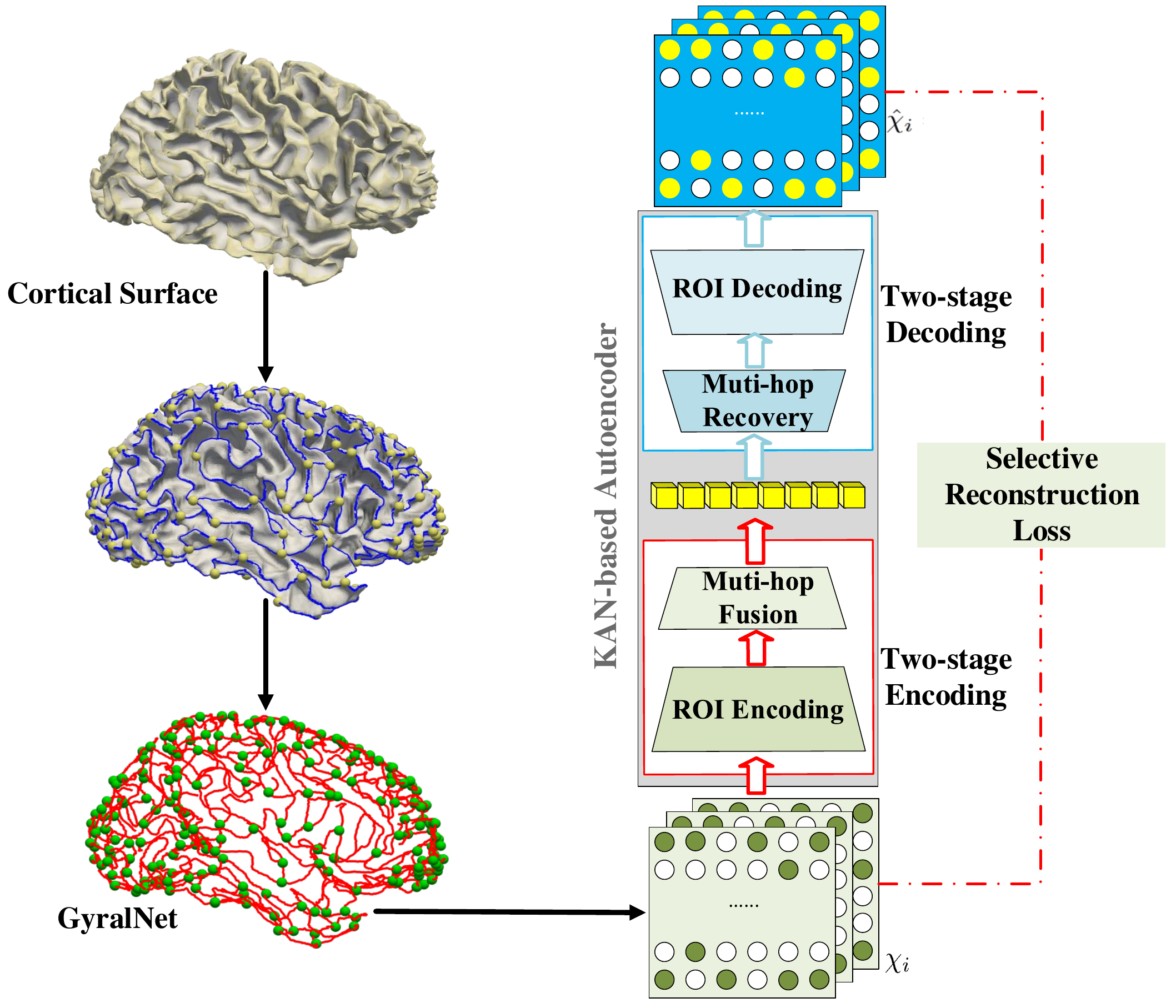}}
\vspace{-0.2cm}
\caption{Overall architecture of the proposed self-supervised framework. The framework constructs a structural similarity enhanced multi-hop feature encoding strategy and uses recent proposed Kolmogorov-Arnold network for anatomical feature embedding of the 3HGs.}
\label{fig:overall}
\end{figure}

\section{method}
\subsection{Data acquisition and preprocessing}
In this study, we use  structural MRI (T1-weighted) of 1064 subjects from Human Connectome Project (HCP) S1200 release~\cite{van2013wu}.
And we applied the same pre-processing procedures as~\cite{zhang2022predicting} for T1 imaging data.
In brief, the pre-processing steps involved brain skull removal, tissue segmentation, and cortical surface reconstruction using the FreeSurfer package. The Destrieux Atlas was then applied for ROI labeling on the reconstructed white matter surface.
The 3HGs are extracted from the cortical surface using the pipeline described in~\cite{chen2017gyral}.
And the undirected graph formed by the topological connections between 3HGs on the gyrus is referred to as \textit{GyralNet}.
\subsection{Hierarchical Multi-hop Encoding with Structural Similarity}
Let $G = (V, E)$ denote the undirected, unweighted graph (GyralNet) with the vertex set $V$ and edges set $E$, where $n = \vert V\vert$ denotes the number of nodes (3HGs) in the graph.
According to~\cite{ribeiro2017struc2vec}, 
the structural similarity between two nodes is determined by comparing the topological structure of the nodes and their neighbors, \textit{i.e.} the degree of the nodes and their neighbors.
When only considering the local topological structure formed by a node and its directly connected neighbors, the feature representation is limited. To improve its expressive power, we adopt a hierarchical representation format to integrate the features and information of  multi-hop neighboring nodes.
The  structural similarity enhanced hierarchical $l$-hop ($l>0$) feature of $i$-th 3HG node can be expressed as:
\begin{equation}\label{eq5}
\chi_i=
\left[
\begin{array}{c}
    \langle S_0\odot I F\rangle _{i} \\
    \langle S_1\odot A_1 F \rangle_{i} \\
     ......\\
     \langle S_{l}\odot A_{l} F \rangle _{i}
\end{array}
\right]
\in \mathbb{R}^{(l+1)\times 75}
\end{equation}
where $F \in \mathbb{R}^{n\times 75}$ is the one-hot ROI label of each 3HGs obtained from Destrieux altas, $A_k$  represents the $k$-hop adjacency matrix of GyralNet , $\langle \cdot \rangle _{i}$ denotes the i-row of the matrix and $\odot$ indicates the Hadamard product. 
$S_k$ represents $k$-hop structural similarity, which is used to measure the similarity between two 3HGs, 
e.g., $u$ and 
$v$:
\begin{equation}\label{eq1}
S_k (u,v)=e^{-w_k(u,v)}
\end{equation}
\begin{equation}\label{eq2}
w_k (u,v)=\mathcal{D}(p(g_k(u)),p(g_k(v)))
\end{equation}
where $g_k(u)$ is the $k$-hop neighborhood set of $u$, $p(\cdot)$ operator denotes generate a ordered degree sequence of a set of nodes and $\mathcal{D}(\cdot)$ measures the similarity between these two ordered degree. 
Note that since the number of k-hop nodes for 
$u$ and  $v$ may not be equal, thus directly calculating the Euclidean distance between sequences cannot accurately reflect their similarity.
Instead, Dynamic Time Warping algorithm (DTW)~\cite{muller2007dynamic}, commonly used in time series analysis to find the optimal alignment between two sequences $M$ and $N$, is employed to measure similarity between neighborhoods of two 3HGs. For each element $a\in M$ and $b \in N$, the cost function we use for DTW in this work is shown in Eq.~\ref{eq3}.
For DTW, we construct a cumulative distance matrix $C \in \mathbb{R}^{\vert M\vert\times \vert N\vert}$, where $C(i,j)$ represents the minimum cumulative distance to $(i,j)$.
The recursive process of $C$ is shown in Eq.~\ref{eq4}, where $C(1,1)=d(1,1)$ and the minimal cumulative distance between the sequences $A$ and $B $ is given by $C(I,J)$. 
This minimum cumulative distance is used to represent the similarity between the two sequences.
\begin{equation}\label{eq3}
d(a,b)=\frac{max(a,b)}{min(a,b)}-1 \in [0,n-1]
\end{equation}
\begin{equation}\label{eq4}
C(i,j)=d(i,j)+min(C(i-1,j),C(i-1,j-1),C(i,j-1))
\end{equation}

\subsection{Self-supervised Embedding Framework }
As shown in Fig.~\ref{fig:overall}, the self-supervised framework for anatomical feature embedding in the 3HG employs a two-stage encoding process $\Phi$ that hierarchically maps structural-similarity-enhanced multi-hop features to a latent representation $\delta$ and a two-stage decoding process $\hat{\Phi}$, which hierarchically reconstructs the original input from this latent representation.
Each encoding and decoding process is implemented by a KAN layer, and the self-supervised framework $f$ is a combination of four layers using the projection operator $\circ$:
\begin{equation}\label{eq6}
\hat{\chi}_i=f(\chi_i)=(\hat{\Phi}_{ROI}\circ\hat{\Phi}_{MH}\circ\Phi_{MH}\circ\Phi_{ROI})\chi_i
\end{equation}
where $\Phi_{ROI}$ is the first stage of encoding, which maps the one-hot ROI encoding into a low-dimensional embedding $\theta_i$, while $\Phi_{MH}$ fuses multi-hop embeddings into a single embedding vector $y_i$.
The decoding process is symmetrical with the encoding process, and consists of two steps: multi-hop feature recovery $\hat{\Phi}_{MH}$ and ROI feature recovery $\hat{\Phi}_{ROI}$.

\noindent\textbf{Loss function.} 
In \textit{cortex2vec}~\cite{zhang2023cortex2vector}, we directly use Mean Square Error (MSE) as the objective function for self-supervised training process:
\begin{equation}\label{eq7}
\mathcal{L}_{mse}=\Vert \hat{\chi_i}- \chi_i  \Vert_2 + \Vert \hat{\theta_i}- \theta_i  \Vert_2
\end{equation}
However, as discussed in~\cite{wang2016structural}, due to the sparsity of GyralNet structure, using MSE as the loss function would make it easier to reconstruct the zero elements in matrix, which is not our intention. Therefore, we need to increase the penalty for reconstruction errors on the non-zero elements within the loss function.
\begin{equation}\label{eq8}
\mathcal{L}=\Vert (\hat{\chi_i}- \chi_i)\odot(\mathbbm{1}_\chi+\lambda \chi_i)  \Vert_2 + \Vert (\hat{\theta_i}- \theta_i)\odot(\mathbbm{1}_\theta+\lambda \theta_i)   \Vert_2
\end{equation}
The revised loss function named as the selective reconstruction loss is shown in Eq.~\ref{eq8}, where $\mathbbm{1}_\chi$ and $\mathbbm{1}_\theta$ are matrices with all elements equal to 1 and $\lambda$ is a hyperparameter.

\section{Experiments}
\subsection{Experiment Settings}
\noindent\textbf{Dataset.} Using GyralNet~\cite{chen2017gyral}, 338,555 3HG points are identified from 1064 subjects in the HCP dataset.
Following the settings in~\cite{zhang2023cortex2vector}, we randomly selected 864 individuals for training, 78 individuals for validation, and 122 individuals for testing.
Each 3HG is treated as a data sample.
For the structural similarity enhanced multi-hop features $\chi_i$, we generated 1-hop features ($l = 1$ in Eq.~\ref{eq5}), 2-hop features ($l = 2$) and 3-hop features ($l = 3$). 
For each type of feature, we trained the model to learn the corresponding ROI embeddings.

\noindent\textbf{Implementation details.}
The KAN-based self-supervised framework is implemented through an open-source library~\footnote{https://github.com/Blealtan/efficient-kan}  and the parameters are initialized following the Xavier scheme.
As for the hyperparameter settings, we set the dimension of the vector $\delta$ in the latent space to 128 and $\lambda$ in Eq.~\ref{eq8} to 2.
We utilize the Adam optimizer with learning rate of 1e-4,  decay rates of 0.5 and 0.999 for the first and second moment estimates, respectively.
We utilize the ReduceLROnPlateau scheduler to facilitate rapid convergence. 
The learning rate reduction is triggered if no improvement in the monitored metric is observed over 10 consecutive epochs.
Our experiments were all conducted on a PC with a NVIDIA TITAN RTX GPU and  a 3.6-GHz Intel Core i7 processor.
\begin{figure}[htb]

  \centering
  \centerline{\includegraphics[width=\linewidth]{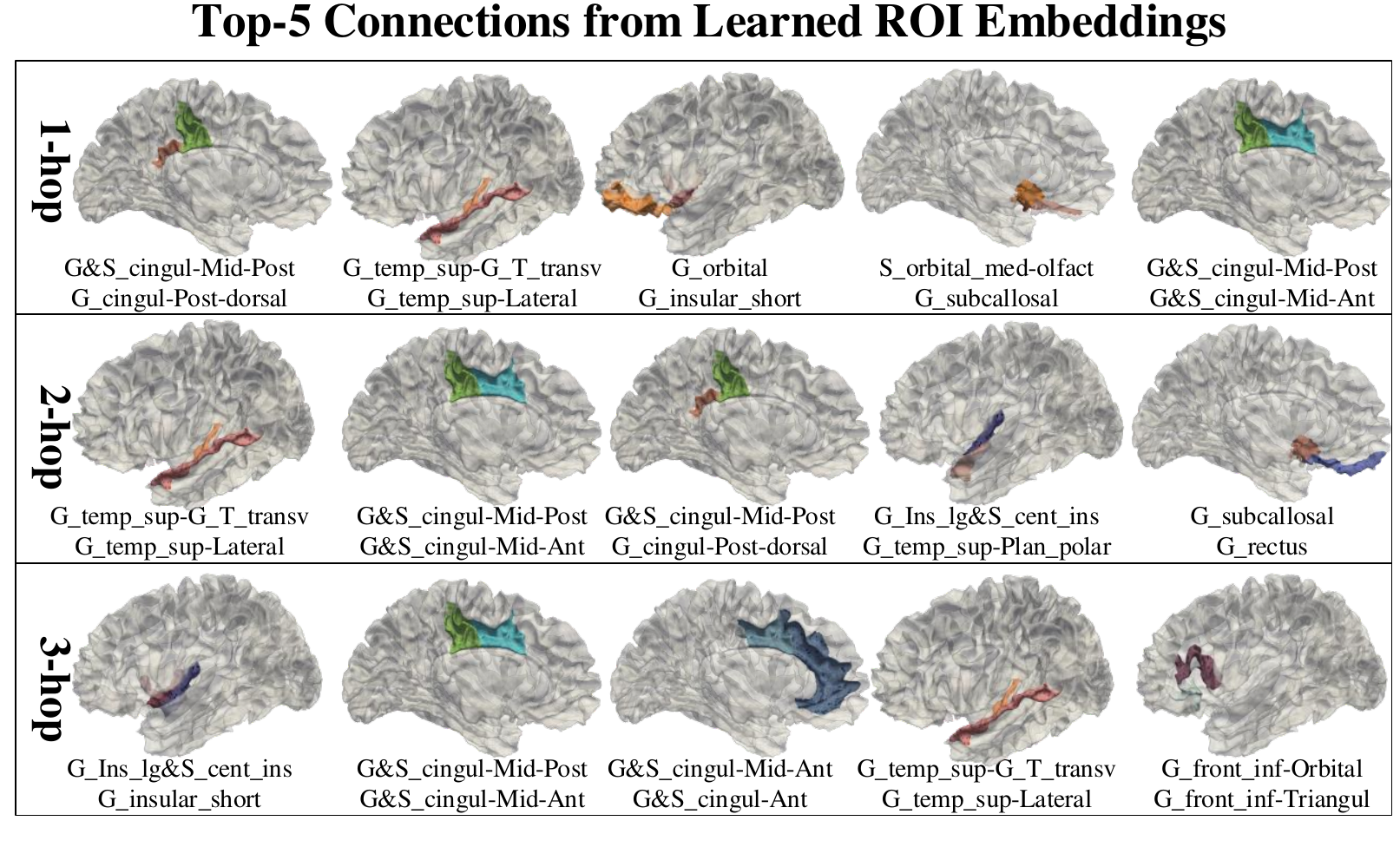}}

 \vspace{-0.4cm}
\caption{\textbf{ROI embedding connectivity.} The Top-5 connections (PCC between pairs of ROIs) from the learned ROI embeddings $\Phi_{ROI}$ are derived using 1-hop, 2-hop, and 3-hop features.
The division of ROIs is based on the Destrieux Atlas.
Within each row, connections are ordered by their rank from left to right.
Color lookup table provided by Freesurfer.}
\label{fig:connctivity}
\end{figure}

\subsection{Results}
\begin{figure}[htb]

  \centering
  \centerline{\includegraphics[width=\linewidth]{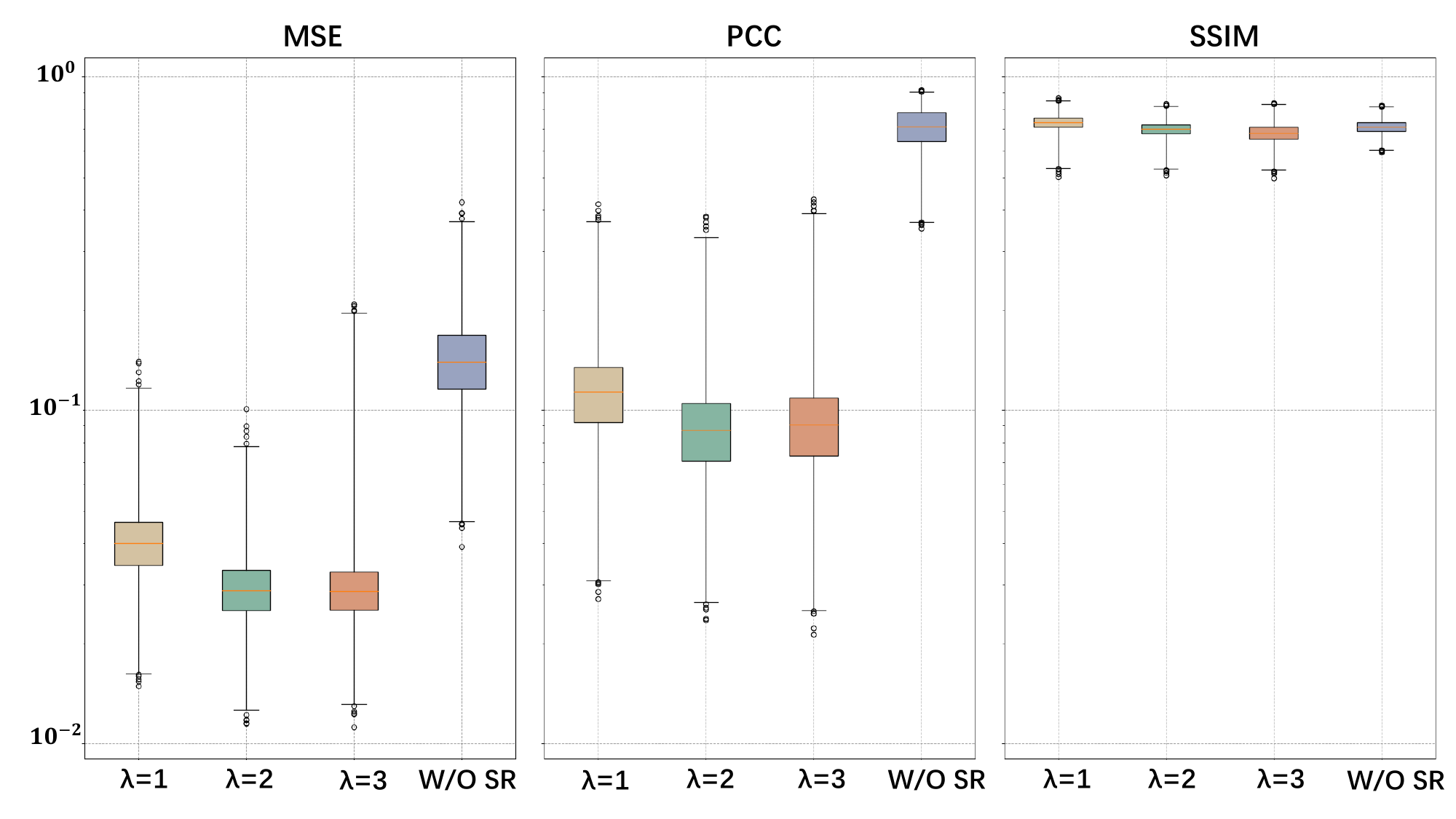}}

\vspace{-0.5cm}
\caption{\textbf{Quantitative evaluation.} Evaluation of different hyperparameter setting and ablation analysis of the proposed selective reconstruction loss (w/o SR) via MSE, PCC and SSIM.
Note that the y-axis is on a log-scale.
}
\label{fig:boxplot}
\end{figure}

As shown in Fig.~\ref{fig:connctivity}, we calculated the Pearson correlation coefficient (PCC) between the learned parameters $\Phi_{ROI}$ for each individual ROI embedding and listed the Top-5 with the strongest connectivity.
Compared with the results reported in~\cite{zhang2023cortex2vector}, the strength of the gyrus-sulcus connection in our method is relatively high, which is mainly because the structural similarity enhanced multi-hop features are more sensitive to the connections at the peripheral areas of GyralNet, \textit{i.e.} the sulcal regions.
In Fig.~\ref{fig:boxplot}, we evaluate the impact of the reconstruction hyperparameter $\lambda$ used to penalize non-zero elements in Eq.~\ref{eq8} on the reconstruction performance. In addition, we also provide an ablation study on the effect of training the network directly using MSE loss instead of using the proposed selected reconstruction loss.
By comparing the MSE, PCC, and Structural Similarity Index Measure (SSIM) between the network-reconstructed features and the ground truth, we observed that when setting $\lambda$ to 2, the reconstructed features show the most consistent pattern with the ground truth. 
Additionally, omitting the proposed selective reconstruction loss results in significant performance degradation across all three metrics.
\vspace{-0.3cm}
\vspace{-0.7cm}

\begin{table}[h!]
    \centering
    \caption{\textbf{Quantitative results.} ROI hit rate of the proposed method under several hop number and the baseline methods.}
    \begin{tabular}[width=\linewidth]{c|c|c|c}
      \hline
     \multicolumn{1}{c|}{\multirow{2}{*}{Methods}} &  \multicolumn{3}{|c}{ROI Hit Rate (\%) $\uparrow$} \\
        \cline{2-4}& L. Hemisphere& R. Hemisphere &Total   \\
      \hline
      cortex2vec~\cite{zhang2023cortex2vector}   & 34.43 & 33.91& 34.15\\
      cortex2vec+sse  & 34.58 & 33.80 & 34.17\\
      \hline
      ours (1-hop)   & 34.49 & 34.05 & 34.26 \\
      ours (2-hop) & \textbf{34.70}& 33.82 & 34.24 \\
      ours (3-hop)  & 34.63 & \textbf{34.17} & \textbf{34.39}\\
      \hline
    \end{tabular}

    \label{tab:table1}
\end{table}
\begin{figure}[htb]

  \centering
  \centerline{\includegraphics[width=\linewidth]{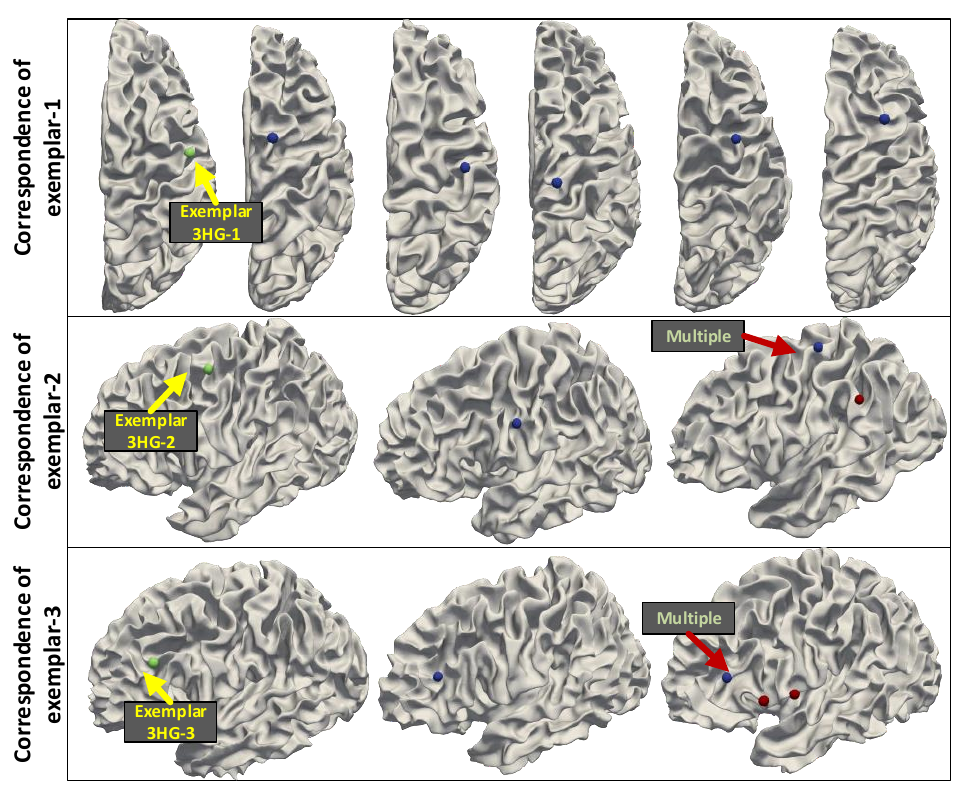}}
\vspace{-0.3cm}
\caption{\textbf{Illustration of cross-subject correspondence.}
All 3HGs are represented as bubbles, with yellow indicating exemplars, blue indicating the identified corresponding points in other subjects, and red indicating the candidate points with cosine similarity close to the corresponding point.}
\label{fig:correspondence}
\end{figure}
We randomly selected a subject as the anchor and established correspondences between each 3HG point on the anchor subject and 38,859 3HG points across 122 subjects in the test set. It is important to note that, due to the variability in brain folding patterns, the number of 3HG nodes varies between individuals. 
Our focus is solely on ensuring that each node on the anchor project has a corresponding node in every individual in the test set, without enforcing a bijective correspondence.
We calculate the cosine similarity of the embeddings produced by the encoder in the proposed framework for the nodes from both the anchor project and the test set. 
The point with the highest similarity is considered the corresponding point to the anchor node, provided the cosine similarity exceeds a threshold (set to 0.9 during experiment). 
Table~\ref{tab:table1} presents the ROI hit rate of the identified corresponding nodes and the anchor nodes, indicating whether they are located in the same brain regions.
The model utilizing feature encoding with 3-hop structural similarity enhancement achieved the best performance. Additionally, when the baseline incorporates the proposed structural similarity enhanced multi-hop feature embeddings (cortex2vec+sse), the hit rate is also improved by nearly $1\%$ relative to the baseline, demonstrating the effectiveness of the proposed encoding strategy.
Figure~\ref{fig:correspondence} shows three examples of 3HG correspondences across individuals. The 3HGs identified in other subjects, corresponding to the anchor subject, are located at similar positions on the cortical surface within their individual space.
These correspondences represent 3HGs with the highest cosine similarity to the anchor 3HG's embedding vector. In most cases ($\approx 70\%$), no other 3HGs exhibit comparable cosine similarity, avoiding ambiguity in the matching process.
3HGs in the frontal and parietal lobe regions exhibit high variability, yet clear one-to-one correspondences can still be identified in these areas, while cortex2vec fails.
However, "competing points"—indicated by red bubbles in the figure—may introduce potential one-to-many correspondences.

\section{Conclusion}
In this paper, we propose a self-supervised auto-encoding framework to embed a finer-scale brain folding pattern -- the 3HG.
The framework leverages structural similarity enhanced-multi-hop features to capture the topological information of anatomical folding patterns and introduces selective reconstruction loss to recover detailed features. Additionally, KAN is employed to improve representation learning capability of the framework.
Extensive experimental results demonstrate that our framework can effectively identify cross-subject correspondences of the 3HGs.
Future work will focus on incorporating additional features from other modalities to enhance cross-subject correspondences and assign adaptive weights to the encoded features during multi-hop feature encoding.
%
%
%
\bibliographystyle{splncs04}
%
\bibliography{mybib}
\end{document}